# Automated Detection of Acute Leukemia using K-mean Clustering Algorithm

[1]Sachin Kumar,  [1]Sumita Mishra, [1]Pallavi Asthana, [2]Pragya

[1]Amity University, Lucknow Campus.
`{skumar3, smishra3, pasthana}@lko.amity.edu`
[2] Department of Chemistry, MVPG College, Lucknow University
`dr.pragya2011@gmail.com`

**Abstract.** Leukemia is a hematologic cancer which develops in blood tissue and triggers rapid production of immature and abnormal shaped white blood cells. Based on statistics it is found that the leukemia is one of the leading causes of death in men and women alike. Microscopic examination of blood sample or bone marrow smear is the most effective technique for diagnosis of leukemia. Pathologists analyze microscopic samples to make diagnostic assessments on the basis of characteristic cell features. Recently, computerized methods for cancer detection have been explored towards minimizing human intervention and providing accurate clinical information. This paper presents an algorithm for automated image based acute leukemia detection systems. The method implemented uses basic enhancement, morphology, filtering and segmenting technique to extract region of interest using k – means clustering algorithm. The proposed algorithm achieved an accuracy of 92.8% and is tested with Nearest Neighbor (kNN) and Naïve Bayes Classifier on the dataset of 60 samples.

**Keywords:** Image Processing, White Blood Cells, Leukemia, Clustering.

## 1    Introduction

Leukemia is a type of cancer which develops in blood tissue. Leukemia originates in the soft inner part of the bones known as Bone Marrow. The Bone Marrow comprises hematopoietic stem cells; which over an interval of period develop into various components of blood namely White Blood Cells (WBC), platelets and Red Blood Cells (RBC), each of them has different roles to play [1].  Leukemia affects the production of White blood cells and interrupts normal cell activities. In addition, Leukemic cells have abnormal growth and their survival time is much more than the normal cells, consequently number of abnormal blood WBC increases rapidly in the blood. Classification of Leukemia is based upon how fast it becomes severe and it can be classified as chronic or acute. Acute leukemia progresses rapidly, and if not treated, becomes fatal within a few months. Chronic Leukemia grows over a longer interval of time, and patients in most cases can live for many years. But chronic leukemia is generally harder to cure than acute leukemia [2 -4]. It can be sub-categorized as Acute Lymphocytic Leukemia (ALL), Acute Myeloid Leukemia (AML), Chronic Lymphocytic Leukemia (CLL) and Chronic Myeloid Leukemia (CML) .This paper deals with automated detection of Acute Lymphocytic Leukemia (ALL). Onset of Acute Lymphocytic Leukemia (ALL) causes production of immature lymphocytes in excessive amounts in bone marrow. These abnormal lymphocytes which are in excess will interrupt the function of normal cells.

In a conventional setup, the pathologist plays a crucial role in accurate diagnosis of ALL; since manual detection process is tedious, time consuming and accuracy of diagnosis also depends upon the experience of pathologist. Recent advances in digital Image processing technology has led to a lot of research towards the development of automated recognition systems for identification of ALL [5-7]. CAD systems have the potential to provide valuable assistance to the pathologist in determination of the presence or the absence of the disease. In addition, it may also help in evaluation of stage of progression of disease.

Several Algorithms of identification and detection of Leukemia have been implemented, S.Mohapatra et al in 2012 [8] segregated region of interest using color based clustering, they achieved successful classification of infected cells using SVM classifier based on Fractal dimensions, shape and color. Later in 2012[9], the same classification was achieved using ANN classifier, features in consideration included algorithms based on color methods. Nasir et al in 2013 [10] did classification of Acute Leukemia cells using Multilayer Perceptron and simplified Fuzz ARTMAP neural networks with FNN and Bayesian classifier, features used for identification were based on shape and color of the target. In 2014, N.Chatap et al [11] did analysis of blood samples for counting Leukemia cells using SVM and KNN nearest neighbor algorithm based on shape of the cells. Later in 2015, R devi et al [12] worked on the classification of Acute Myelogenous Leukemia in Blood Microscopic Images using PNN considering features based on shape, similar work was done by L. Faivdullah et al [13] for Leukemia detection from blood smears using SVM. The proposed work segregates the region of interest using K-means clustering, and then a combination of morphological, color, Geometric, Textural and statistical features has been used to classify a mature lymphocyte and leukemic lymphocyte using Nearest Neighbor (kNN) and Naïve Bayes Classifier. Further devised methodology also addresses the segmentation of overlapping cells

The paper is arranged in four sections. The next section deals with the proposed methodology and algorithm. Section III deals with identification and classification, and section IV provides the concluding remark.

## 2    Proposed Methodology

The automated detection of Leukemia is performed by analyzing numerous microscopic images of the white blood cells or bone marrow smears obtained from the patient. First step is image acquisition; the samples for the proposed work were obtained from Dr RML Awadh hospital, Lucknow. Proposed algorithm for the automated identification of Acute Leukemia from microscopic image is shown in Figure 1. Various steps involved are discussed below:

### 2.1    Image Preprocessing

The images of acceptable quality are subjected to pre-processing operations. The acquired microscopic digital image is usually corrupted by various kinds of noise or it may have a blurred region in the image which is important for detection. Figure 2 depicts the microscopic image of one of the blood samples.

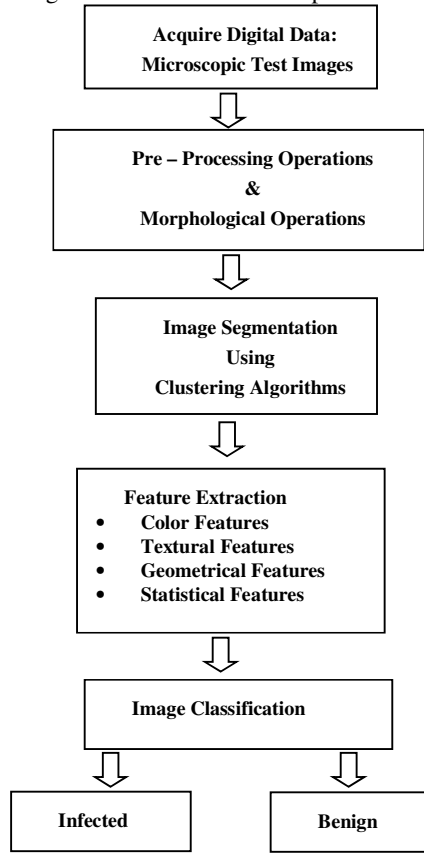

**Fig. 1.** Proposed algorithm for automated identification of leukemia

Image preprocessing operations are performed to suppress the undesired distortions present in the image and enhance image features relevant for further analysis. The noise is removed from the image using various filtering techniques depending upon the type of noise. We have utilized Wiener filter which adequately reduced the blurriness without reducing the image sharpness. Further histogram equalization technique is applied to enhance the contrast in the image. Figure 3 depicts histogram of the image for performing adaptive histogram equalization. Gray-scale transformation Brightness thresholding is chosen to modify brightness and threshold used is 192. This gray-scale transformation results in a binary image as shown in Figure 4. [14-16].

The component mainly analyzed in the dataset is leucocyte other than that every other component needs to be eliminated from the dataset. Further In the dataset being examined it is possible that certain percentage of leucocytes is present on the edges of the image. In the image cleaning process, the leucocytes which are at the edge in the sample image under study and other irrelevant elements present in the image are removed in order to reduce errors in the later stages of the identification process. There are two cleaning operations which are required
- Removal of Noise or Cleaning of Image
- Removal of abnormal components

The first one is done with help of filtering algorithms especially wiener and median filter. The component having a small area is usually the component located on the edges. The component with large values of area must be cells that are overlapping leucocytes. Therefore area and the convex area both need to be calculated for the removal of the unwanted components. Figure 5 depicts the morphologically cleaned image which is obtained by removing the leucocytes on the edges and irregular components.

2.2    Image Segmentation.

Segmentation process partitions an image into distinct regions on the basis of features of interest. Segmentation in the present work involves segregation of white blood cells. Five components of white blood cells include: Neutrophil, Basophil, Eosinophil, Lymphocyte and Monocyte. ALL symptoms are associated only with the lymphocytes since morphological components of normal and malignant lymphocytes are significantly different ; so other four components of white blood cells namely neutrophil, basophil, eosinophil and myelocytes, are neglected during segmentation process.

Figure 6, Figure 7 and Figure 8 depicts results of the k – means clustering with segregation of different nucleus and cytoplasm. [17-19].The performance of segmentation approaches such as k – means [20], texture based segmentation [21] and color based segmentation [22] have been compared. The brief description of the performance measures used is:

- Probability Random Index (PRI): It is a nonparametric evaluation of the goodness of segmentation. It is obtained by summing the number of pixel pairs with same label and number with different label in both S (test samples) and G (ground reali-ty) and then dividing it by total number of pixel pairs. For a given a set of ground truth segmentations $G_k$, PRI is evaluated using:

$$\text{PRI}(S_{test}, G_k) = \frac{1}{(N/2)} \sum_{\forall i,j \& i<j} [c_{ij}p_{ij} + (1 - c_{ij})(1 - p_{ij})] \quad (1)$$

Where $c_{ij}$ is an event that describes a pixel pair (i, j) having same or different label in the test image $S_{test}$

- Variance of Information (VOI): Variation of Information gives the measure of distance between two clusters. It gives partition of pixels with different clusters. Clustering with clusters is represented by a random variable X, X = {1… k} such that $P_i = \frac{|X_i|}{n}, i \in X, and\ n = \sum_i X_i$ is the variation of information between two clusters X and Y. Thus VOI(X, Y) is represented using

$$\text{VOI}(X, Y) = H(X) = H(Y) - 2I(X, Y) \quad (2)$$

Where H(X) is entropy of X and I(X, Y) is mutual information between X and Y.

- Global Consistency Error (GCE): Local refinement error is calculated using equation (3), where $s_i$ and $g_j$ contain pixel, $p_k$, so that s ∈ S, g ∈ G, where S segment is obtained after segmentation by the algorithm being evaluated and G denotes reference segment. The value obtained from (3) is used to evaluate global consistency errors using (4), where n denotes set of difference operation. R(x, y) represents the set of pixels corresponding to region x that includes pixel y. GCE quantify the amount of error in segmentation. Table 1 depicts the comparative values of segmentation algorithms on the basis of VOI, PRI and GCE for the dataset evaluated. [23-25]

$$E(s_{i,}g_{j,}p_k) = \frac{|R(s_i,p_k)/R(g_j,p_k)|}{|R(s_i,p_k)|} \quad (3)$$

$$\text{GCE}(S, G) = \frac{1}{n} \min\{ \sum E(S, G. p_i), E(S, G. p_i)\} \quad (4)$$

**Table 1.** Qualitative Analysis of Performance Parameters

|              | PRI   | GCE    | VOI   |
|--------------|-------|--------|-------|
| Color k-means | 0.932 | O.0079 | 0.089 |
| k-means      | 0.942 | 0.0091 | 0.092 |
| Texture based | 0.941 | 0.0132 | 0.015 |

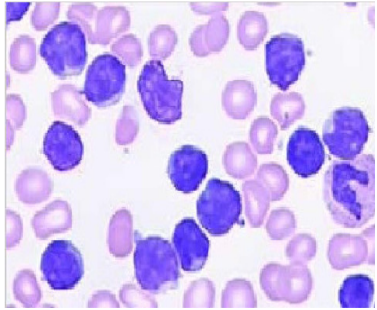

**Fig. 2.** Microscopic Image of blood Samples

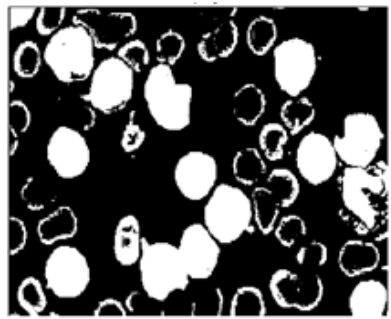

**Fig. 4.** Binary Image

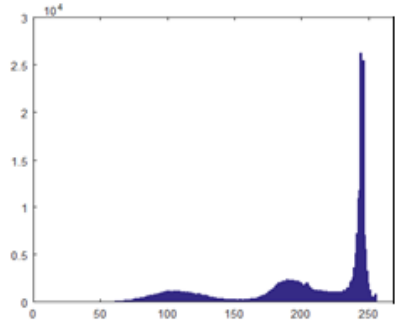

**Fig. 3.** Histogram for Adaptive Thresholding

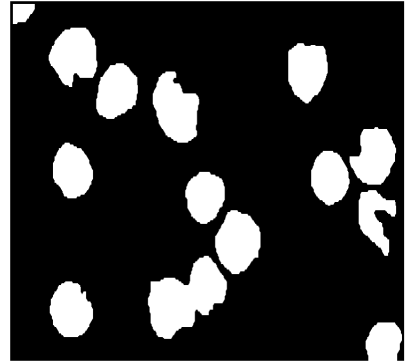

**Fig. 5.** Morphologically Cleaned Image

## 3 Identification and Classification

Identification involves extraction of color, Geometric, Textural and statistical features. Final step i.e. the labeling of sample as malignant or benign is achieved through image classification process. Image classification analyzes various image features to arrange data into categories. Classification algorithms typically employ two processes: training and testing. In the training process, relevant properties of typical image features are isolated and a unique description of each classification category is created.

Two categories of classification algorithms namely supervised and unsupervised are generally used. In supervised classification, statistical processes are employed to extract class descriptors. Classification used in the present work relies on clustering algorithms to automatically segment the training data into various prototype classes. During the testing phase, features of sample dataset are compared with the previously calculated standard values. Depending upon the values of the input image finally clas-sification is achieved with the help of Nearest Neighbor (kNN) and Naïve Bayes Classifier, comparison of which is also presented. The work was carried out on the dataset of 60 samples.

**3.1** Identification of grouped leucocytes.

Microscopic images of blood samples usually contain cells which are overlapping, this complicates the analysis and identification process. Segregation of Region of Interest (ROI) is achieved through k-Means clustering. In order to segregate leucocytes roundness has been used as a measure. Roundness checks whether the shape of the object is circular or not by excluding the local irregularities. Roundness can be gained by dividing the area of a circle to the area of an object by using the convex perimeter.

$$Roundness = \frac{4 \; X \; \pi \; X \; area}{convex\_perimeter^2} \quad (5)$$

The value of roundness is 1 if the object is circular and the value of roundness is less than 1 for the non-circular objects. Roundness as a measure is less sensitive to irregular boundaries because it excludes the local irregularities. Threshold value chosen is 0.80 to distinguish between the single leucocyte and clusters of overlapping leucocytes. The components which are having the roundness value more than the value of threshold are considered as the individual leucocyte while the components which are having value less than the threshold are considered as grouped leucocytes. Figure 9 represents roundness metrics obtained for various leucocytes. The individual leucocytes are sent next for the further study and the grouped leucocytes are sent to the separation process. [26-28]

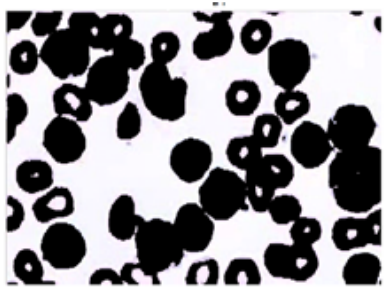
**Fig. 6.** Cluster 1: k-means Clustering

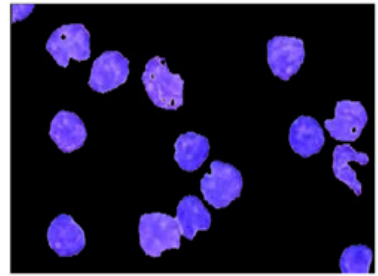
**Fig. 7.** Cluster 2: k-means Clustering

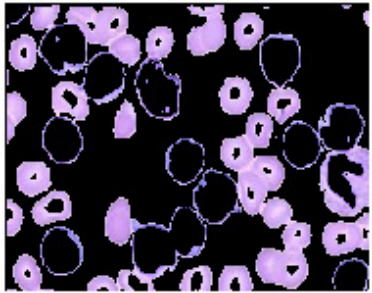
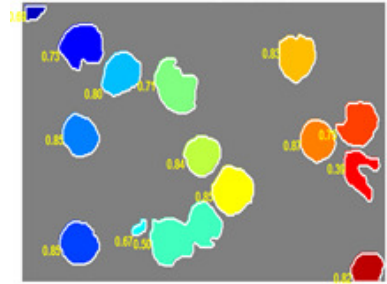

**Fig. 8.** Cluster 3: k-means Clustering

**Fig. 9.** Metrics Close to One Indicates Roundness.

Solidity is used to find out the density of a component. It is obtained as:

$$Solidity = \frac{area}{convex\ area} \quad (6)$$

If the value is 1 then it can be identified as a solid object. If the value is less than 1 then it is a component having irregular boundaries. The threshold value for solidity which is used for identifying the abnormal components is obtained from the image which is having individual leucocytes. An optimum threshold value of 0.80 can efficiently be used to find out abnormal components from the image this is depicted in Figure 10. The components which are having the solidity value less than the threshold are removed. [29, 30]

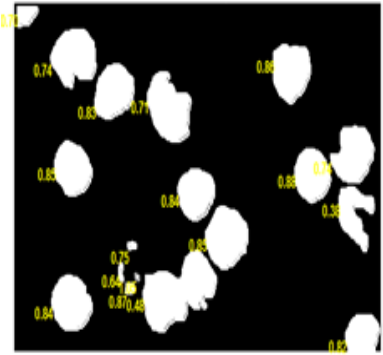
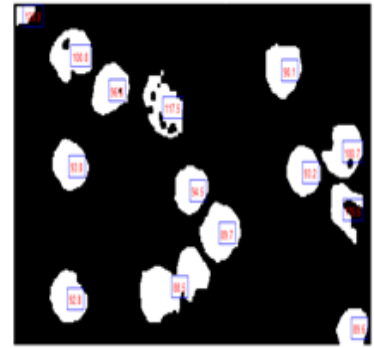

**Fig. 10.** Measure of Solidity

**Fig. 11.** Standard Deviation of identified leucocytes

### 3.2 Nucleus and cytoplasm selection

The leucocytes segmented can now be used to extract the nucleus and cytoplasm. This is achieved by cropping the image with the bounding box. This step separates out each leucocyte. The borders of images obtained in the above step have to be cleaned in order to proceed further. Next step involves cropping out the outer portion of the leucocyte to segregate cytoplasm. This process segregates cytoplasm. From the close examination it can be concluded that white blood cells nuclei are more in contrast on the green component of the RGB color space. So, we can get nucleus by using the threshold. [31]

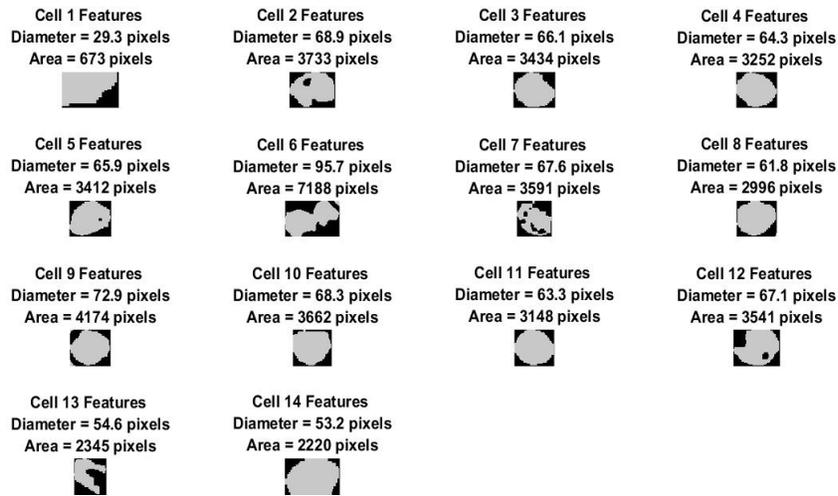

**Fig. 12.** Segmented leucocytes with Corresponding Diameter and Area

### 3.3 Feature Extraction

Feature extraction is the process of converting the image into data so that we can check these values with the standard values and finally identify infected samples. Figure 12 depicts individual segmentation of the lymphocytes with its area and diameter. Features required to train model parameters include are:

- Color Features – It includes mean color values of the grey images acquired.
- Geometric Features – It includes perimeter, radius, area, rectangularity, compactness, convexity, concavity, symmetry, elongation, eccentricity, solidity etc.
- Texture Features – It Includes entropy, energy, homogeneity, correlation as obtained.

- Statistical Features – It includes mean, variance and standard deviation. The values are computed are shown in Figure11.

$$Elongation = 1 - \frac{major\ axis}{minor\ axis} \quad (7)$$

$$Eccentricuty = \frac{\sqrt{(major\ axis^2 - minor\ axis^2)}}{major\ axis} \quad (8)$$

$$Rectangularity = \frac{area}{major\ axis\ X\ minor\ axis} \quad (9)$$

$$Convexity = \frac{Perimeter\_convex}{Perimeter} \quad (10)$$

$$Compactness = \frac{4\ X\ \pi\ X\ area}{perimeter^2} \quad (11)$$

Elongation indicates the object elongation towards particular axis. Rectangularity depicts how well the bounding box is filled. Eccentricity is the ratio of the major axis length and the foci of the ellipse. Convexity shows the relative amount of difference of object from its convex object. Compactness is the ratio of the area of an object and area of circle having same perimeter. Figure 13 depicts the geometric characteristics of the infected cell and Figure 14 depicts texture features of the infected cell. [32, 33]

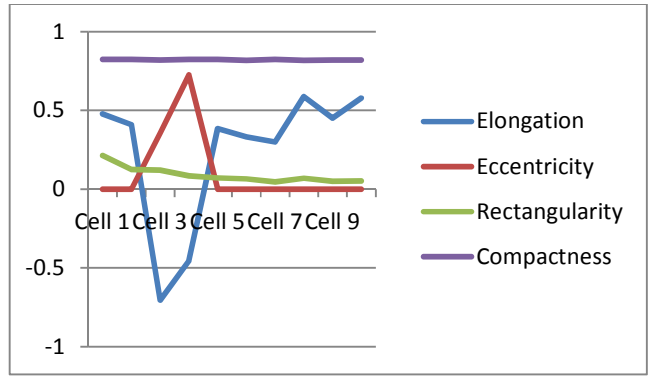

**Fig. 13.** Geometric Features of Infected Cells.

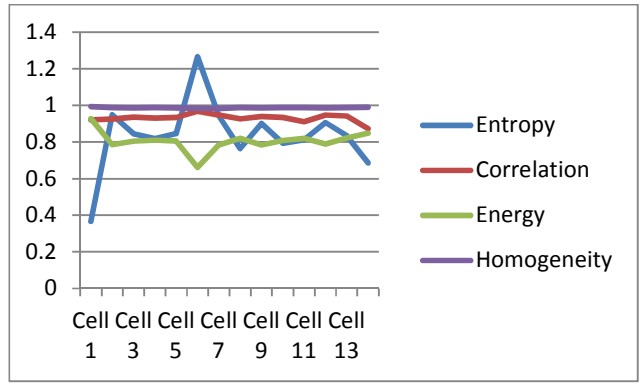

**Fig. 14.** Texture Features of Infected Cells.

### 3.4 Image Classification

Proposed algorithm is tested with Nearest Neighbor (kNN) and Naïve Bayes Classifier on the dataset of 60 pretested samples, the accuracy achieved is 92.8%.
The process involved following steps

1. Finalization of feature set
2. Selection of appropriate Algorithm
3. Mapping and Training of model parameters

Features such as Elongation, Eccentricity, Rectangularity, Convexity, Compactness, entropy, energy, homogeneity, correlation and standard deviation are used to train model parameters to identify infected cells. Confusion Matrix has been utilized to compute performance of classifiers. Performance parameters included accuracy, sensitivity and specificity shown in Figure 15.

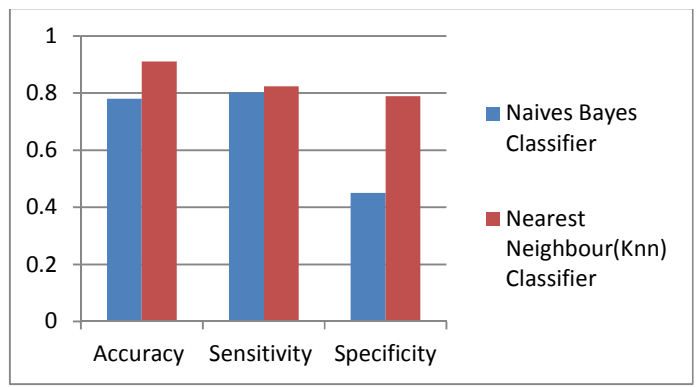

**Fig. 15.** Performance Analysis of classifiers

## 4 Conclusion

The chances of successful outcome of cancer treatment are primarily dependent on its early detection and diagnosis. Acute Lymphocytic Leukemia is the most common form of blood cancer which can identified through examination of blood and bone marrow smears by trained experts. This Manual inspection process is time-consuming and error prone, thus a computer-based system for automated detection of Acute Lymphocytic Leukemia may provide an assistive diagnostic tool for pathologists. The automated segregation and identification algorithm aims to reduce the latency period involved in treatment, which is sometimes life threatening. The proposed automated system is tested with Nearest Neighbor (kNN) and Naïve Bayes Classifier on the dataset of 60 pretested samples, the accuracy achieved is 92.8%. The results show that algorithm proposed achieves an acceptable performance for the diagnosis of Acute Lymphocytic Leukemia; further the devised methodology also addresses the segmentation of overlapping cells.